\documentclass[preprint,12pt]{elsarticle}

\usepackage{amssymb}
\usepackage{epsfig}
\usepackage{epsf}
\usepackage{booktabs}
\usepackage{amssymb}
\usepackage{multirow}
\usepackage{mathrsfs}
\usepackage{longtable}
\usepackage{lscape}
\usepackage{tabularx}
\usepackage{subfigure}
\newdefinition{definition}{Definition}
\newdefinition{example}{Example}
\newdefinition{remark}{Remark}
\newdefinition{property}{Property}

\journal{Artificial Intelligence}

\begin{document}

\begin{frontmatter}



\title{Generalized Evidence Theory}

\author[SWU,VU,SJTU]{Yong Deng\corref{COR}}
\ead{ydeng@swu.edu.cn; prof.deng@hotmail.com}
\cortext[COR]{Corresponding author: Yong Deng, School of Computer and Information Science, Southwest University, Chongqing 400715, China.}

\address[SWU]{School of Computer and Information Science, Southwest University, Chongqing, 400715, China}
\address[VU]{School of Engineering, Vanderbilt University, Nashville, TN, 37235, USA}
\address[SJTU]{School of Electronics and Information Technology, Shanghai Jiao Tong University, Shanghai 200240, China}

\begin{abstract}
Conflict management is still an open issue in the application of Dempster Shafer evidence theory. A lot of works have been presented to address this issue. In this paper, a new theory, called as generalized evidence theory (GET), is proposed. Compared with existing methods, GET assumes that the general situation is in open world due to the uncertainty and incomplete knowledge. The conflicting evidence is handled under the framework of GET. It is shown that the new theory can explain and deal with the conflicting evidence in a more reasonable way.
\end{abstract}

\begin{keyword}
Dempster-Shafer Evidence Theory \sep Generalized Evidence Theory \sep Belief Function \sep Conflict management \sep Open World \sep Close World
\end{keyword}
\end{frontmatter}

\section{Introduction}

Handling uncertainty is heavily studied in many real applications such as approximate reasoning and decision making. The first step to handle uncertain information is reasonable description or modelling of the uncertain information. One of the most used theory is Dempster Shafer evidence theory. Since first proposed by Dempster\cite{Dempster1967} and then developed by Shafer\cite{Shafer1976}, has been paid much attentions for a long time and continually attracted growing interests\cite{yang1994evidential,yager1996aggregation,denoeux2004evclus,elouedi2004assessing,dempster2006dempster,yang2006evidential,yang2006application,jousselme2006measuring,cuzzolin2007two,cuzzolin2008geometric,klir2008remarks,dempster2008generalization,utkin2009computing,dezert2012hierarchical,
denoeux2012evidential,kang2012evidential,Wei20132564,yager2012dempster,masson2011ensemble,zhang2012assessment,deng2011risk,deng2011modeling,deng2011new,nguyen2012belief,Huang2013,liu2013evidential,denoeux2013maximum,liu2014credal}.

One open issue of evidence theory is the conflict management when evidence highly conflicts with each other. A famous example is illustrated by Zadeh\cite{zadeh1986simple}. Since then, hundred of methods are proposed to address this issue  \cite{murphy2000combining,lefevre2002belief,yong2004combining,liu2006analyzing,haenni2005shedding,SebastDesterckeToward2013,schubert2011conflict,yang2013discounted,roquel2012new,lefevre2013preserve,sarabi2013information,yang2013evidential}.

The normalization step is questionable in Dempster combination rule and is deleted to assign the conflict to the whole sets\cite{yager1987dempster}, under the closed-world assumption that the frame of discernment is exhaustive. The transferable belief model (TBM) is presented to represent quantified beliefs based on belief functions\cite{smets1994transferable,smets2005decision}. TBM was constructed by two levels, namely the credal level where beliefs are entertained and quantified by belief functions and the pignistic level where beliefs can be used to make decisions and are quantified by probability functions. Dubois and Prade proposed a new rule of combination which is better adapted and more specific than Yager's rule of combination concerning the assignment of the conflicting mass\cite{dubois1988representation}. Lefevre et.al reviewed the previous works and made a unified model to handle conflict evidence. The main idea of their method is to address that the conflict will be assigned to which hypothesis and how to assign the conflict\cite{lefevre2002belief}. However, to assign the conflict to different hypothesis is questioned by Haenni \cite{haenni2002alternatives}. A model X with method Y has an intuitive result Z. Haenni argued to modify the method Y is not reasonable since that the intuitive result may be caused by model X. Some methods are coincide with Haenni's opinion. For example, Murphy averages the conflict evidence and then combines the averaged evidence itself several times\cite{murphy2000combining}. A weighted average method is proposed by Deng\cite{yong2004combining} with a better convergence ability. In the weighted average method, the evidence distance function presented in \cite{jousselme2001new}, instead of the commonly used conflict coefficient in evidence theory, is used to model the conflict degree. Liu pointed out that the commonly used conflict coefficient in evidence theory is not reasonable to represent the conflict degree between two pieces of evidence\cite{liu2006analyzing}. To address this issue, a two dimension conflict measure, combined with the classical conflict coefficient and pignistic betting distance, is proposed. In addition, the adoptive is discussed. Three cases are proposed according to the two dimension conflict measure.

Generally speaking, there are two main reasons to cause the evidence conflicts with each other. One is the questionable sensor reliability caused by any disturbance or the maintain condition of the equipment. The other is that the system is in open world where our knowledge is not completed. For example, target recognition is widely used in military application. We know that the enemy has three types of jet fighter, namely ${A,B,C}$. As a result, one of the three types will be recognized, given by some collected evidence from different radars. However, we do not know that a new type of jet fighter, namely $D$ is developed secretly. When the target $D$ is flying near the radars, one radar may sent the report that the target is $A$ and the one radar may sent the report that the target is $C$, which is very similar to Zadeh's count-intuitive example.

The rest of this paper is organized as follows. Section \ref{SectDempster} the background knowledge are  brief introduced, including Dempster-Shafer theory, pignistic probability transform and Jousselme's evidence distance proposed. In Section \ref{SectGET}, the proposed generalized evidence theory is presented, mainly including GET, generalized evidence distance, generalized combination rule (GCR) and its application. The discuss of $\phi$, generalized conflict model and its application are presented  in Section \ref{SectAD}. Finally, conclusions are given in Section \ref{SectConclusions}.

\section{Preliminaries}\label{SectDempster}

\subsection{Dempster-Shafer theory}
For completeness of the explanation, a few basic concepts about Dempster-Shafer theory are introduced as follows.

For a finite nonempty set $\Omega  = \{ H_1 ,H_2 , \cdots ,H_N \}$, $\Omega$ is called a frame of discernment (FOD) when satisfying
\begin{equation}\label{exclusiveness}
H_i  \cap H_j  = \emptyset , \quad \forall i,j = \{ 1, \cdots, N\}.
\end{equation}
Let $2^\Omega$ be the set of all subsets of $\Omega$, namely
\begin{equation}
2^\Omega   = \{ A \;\; | \;\; A \subseteq \Omega \}.
\end{equation}
$2^\Omega$ is called the power set of $\Omega$. For a FOD $\Omega$, a mass function is a mapping $m$ from  $2^\Omega$ to $[0,1]$, formally defined by:
\begin{equation}
m: \quad 2^\Omega \to [0,1]
\end{equation}
which satisfies the following condition:
\begin{eqnarray}
\sum\limits_{A \in {2^\Omega }} {m(A) = 1} \\
m(\emptyset ) = 0 \label{bpaemptyset}
\end{eqnarray}

In Dempster-Shafer theory, a mass function is also called a basic probability assignment (BPA). Given a BPA, the belief function $Bel\;:\;2^\Omega   \to [0,1]$ is defined as
\begin{equation}
Bel(A) = \sum\limits_{B \subseteq A} {m(B)}
\end{equation}
The plausibility function $Pl\;:\;2^\Omega   \to [0,1]$ is defined as
\begin{equation}
Pl(A) = 1 - Bel(\bar A) = \sum\limits_{B \cap A \ne \emptyset }
{m(B)}
\end{equation}
where $\bar A = \Omega  - A$. These functions $Bel$ and $Pl$ express the lower bound and upper bound in which subset $A$ has been supported, respectively.

Given two independent BPAs $m_1$ and $m_2$, Dempster's rule of combination, denoted by $m = m_1 \oplus m_2$, is used to combine them and it is defined as follows
\begin{equation}\label{Dempstercombination}
m(A) = \left\{ {\begin{array}{*{20}l}
   {\frac{1}{{1 - K}}\sum\limits_{B \cap C = A} {m_1 (B)m_2 (C)} \;,} & {A \ne \emptyset ;}  \\
   {0\;,} & {A = \emptyset }.  \\
\end{array}} \right.
\end{equation}
with
\begin{equation}\label{DempsterK}
K = \sum\limits_{B \cap C = \emptyset } {m_1 (B)m_2 (C)}
\end{equation}
Note that the Dempster's rule of combination is only applicable to such two BPAs which satisfy the condition $K < 1$.

\subsection{Pignistic Probability Transform}
In the transferable belief model(TBM)\cite{smets2005decision}, pignistic probabilities are used for decision making.

Let $m$ be a BPA on the frame of discernment  $\Theta$. Its associated pignistic probability  function $BetP_m :\Theta  \to [0,1]$ is defined as
\begin{equation}\label{betp}
BetP_m (\omega) = \sum\limits_{A \subseteq P(\Theta ),\omega \in A} {\frac{1}{{|A|}}} \frac{{m(A)}}{{1 - m(\emptyset )}},\quad m(\emptyset ) \ne 1
\end{equation}

Where ${|A|}$ is the cardinality of subset $A$. The process of pignistic probability transform (PPT) is that basic probability assignment transferred to probability distribution. Therefore, the pignistic betting distance\cite{Smets1994191} can be easily obtained by PPT.
  Let $m_1$ and $m_2$ be two BPAs on the frame of discernment $\Theta$, the pignistic probability functions are $BetP_{m1}$ and $BetP_{m2}$, respectively.
  The pignistic betting distance   $difBetP_{{m_1}}^{{m_2}}$ ($difBetP$, for short) between  $BetP_{m_1}$ and $BetP_{m_2}$ are given as follows.
  \begin{equation}\label{pptdistance}
  difBetP = {\max _{A \subseteq \theta }}(\left| {Bet{P_{{m_1}}}(A) - Bet{P_{{m_2}}}(A)} \right|)
  \end{equation}
 $\left| {Bet{P_{{m_1}}}(A) - Bet{P_{{m_2}}}(A)} \right|$ indicates that the support degree by BPAs.

 \subsection{Jousselme's evidence distance}
Jousselme et al.\cite{jousselme2001new} proposed a new distance of measure the conflicting of two bodies of evidence. That is evidence distance.

Let $m_1$ and $m_2$ be two BPAs defined on the same frame of discernment  $\Theta$, containing $ N $ mutually exclusive and exhaustive hypotheses.
 Let $d_{BPA} (m_1 ,m_2 )$ represents the distance of two bodies of evidence.
\begin{equation} \label{Jousselmedistance}
d_{BPA} (m_1 ,m_2 ) = \sqrt {\frac{1}{2}\left( {{m_1 }  - {m_2 } } \right)^T \underline{\underline D} \left( { {m_1 }  -  {m_2 } } \right)}
\end{equation}

Where $ { {m_1 } } $  and  ${{m_2 } } $ are respective two BPAs. ${\underline{\underline D} } $ is an $2^N \times 2^N $ matrix whose
elements $ \underline{\underline D} (A,B) = \frac{{|A \cap B|}}{{|A \cup B|}} $ with $ A,B \in P(\Theta ) $ from $m_1$ and
$m_2$, respectively.

\subsection{Liu's evidence conflict model}\label{liuconflict}
In traditional Dempster-Shafer evidence, the conflict coefficient $K$ represents the degree of conflicting between two bodies of evidence. Liu pointed out that the traditional conflict coefficient $K$ can not efficiently measure the disagreement between two bodies of evidence\cite{liu2006analyzing}. The conflict model proposed by Liu in \cite{liu2006analyzing} indicates that the pignistic betting distance and coefficient $K$ are united to represent the  degree of conflict.

Let two BPAs $m_1$ and $m_2$ are on the same frame of discernment $\Omega$, the conflict model proposed by Liu is shown as follow.
\begin{equation}
cf({m_1},{m_2}) = \left\langle {K,difBetP} \right\rangle
\end{equation}
Where $K$ is the classic conflict coefficient of Dempster combination rule, and the $difBetP$ is the pignistic betting distance in Eq.(\ref{pptdistance}).
When both $K > \varepsilon$ and $difBetP > \varepsilon$,  $m_1$ and $m_2$ are regard as  conflict, where $\varepsilon  \in [0,1]$ is the threshold of conflict tolerance. Because there does not exist an  "absolute meaningful threshold" of conflict tolerance satisfying all pairs of BPAs\cite{ayoun2001data}, the value of  $\varepsilon$ is subjective and not fixed in different applications. Generally speaking, the value of $\varepsilon$ more closer to 1, the greater the conflict tolerance is.

\section{The Generalized Theory}\label{SectGET}

Dempster-Shafer theory has many merits in information fusion, which is a big improvement, it is relative to the Bayesian theory. However, there are still some drawbacks in Dempster-Shafer theory, including but not only as  follows. Computational complexity is heavy with the increasing elements in the set, which limits the application in the area of  high real-time.  In addition, while the evidence are high conflicting, the counter-intuitive results will present.
 In the frame of discernment, the follow two  may be the main reasons that lead to highly conflicting. One is the incomplete of the frame of  discernment. For example, for military applications, if there are three targets $a$, $b$ and  $c$ on the frame of discernment. Then the sensors can only recognize the different unions of this three targets. However, if there existing  a new unknown target $d$, the sensors can not distinguish whether it is one of the previous three targets. In this situation, the recognition results will be multifarious, after combination, the incorrect results will present. Another is the reliability of sensors itself. Work condition, disturbance, etc. will influence the judgment result.
There are some alternatives to overcome these shortcomings. Preprocessing the information or approximation algorithm\cite{voorbraak1989computationally,haenni2002resource} are used to solve the computational complexity. Researchers have paid great effort to solve the high conflicting problem in recent years. The two typical solution is transferable belief model (TBM)\cite{smets1994transferable} and Dezert-Smarandache theory (DSmT)\cite{dezert2006dsmt}. The characteristic of TBM model is that the interesting concepts of close world and open world. However, to our best knowledge, the application of TBM is still under the close world. DSmT provides a new idea of solution the high conflicting problem, but the computation complexity is heavier, and when under the condition of low conflicting, the result inferior to Dempster-Shafer theory.
In summary, the solution for high conflicting still needs a more reasonable theory model. The new model should be capable of the incomplete of the fame of  discernment, and the computation complexity should not larger than traditional Dempster-Shafer theory. Taking these into consideration, a new evidence theory, called generalized evidence theory (GET) is proposed in this paper.

It should be point out that traditional Dempster-Shafer evidence is based on the frame of discernment, and constraint the BPA $m(\phi)$ must equal to 0. That is the classic  Dempster-Shafer theory can only function in a close world. In this paper, close world means that the elements in the frame of discernment is exhaustive and complete. There are many cases in real-life. For example, the points of a dice only contain the six possibility: 1, 2, 3, 4, 5 and 6. However, in some application, due to lack of complete knowledge, it is possible to obtain a partial frame of discernment, but not a complete frame of discernment. As mentioned above, we know that enemy's targets are "a", "b" and "c", whereas there may exist a secret undeclared target "d". And none of the sensors can recognize the target "d", efficiently. In this situation, the discernment \{a, b, c\} is incomplete. This  is the  situation of open world. Along with the enriched  knowledge, open world is absolute and close world is relative. Another example of SARS, before its appearances, the frame of discernment of pneumonia is sure not complete. Obviously, the traditional Dempster-Shafer evidence can only represent and function uncertain information in the close world, which limits the application of Dempster-Shafer evidence.

\subsection{Basic Concepts of GET}

\begin{definition}\label{defGET}
Supposed $U$ is a frame of discernment of open world, the power set, denoted $2_G^U$, is composed with the $2^U$ propositions, $\forall A \subset U$. A mass function is a mapping, $m: 2_G^U \to [0, 1]$, which satisfies the following conditions:
\begin{equation}\label{GBPA}
\sum\limits_{A \in 2_G^U} {{m_G}(A) = 1}
\end{equation}
Then, $m$ is the generalized basic probability assignment (GBPA) of the frame of discernment $U$. The different between GBPA and traditional BPA is the  Eq.(\ref{bpaemptyset}). Notice that the  $m(\phi)=0$ is not necessary in GBPA. In other words, the empty set can also be a focal element. If $m(\phi)=0$, the GBPA degenerates as a traditional BPA.
\end{definition}

The $\phi$ is used to model open world in GET. It should be emphasized that the $\phi$ in GET is not a common empty set, which means that a focal element or the unions of focal elements that out of the given frame of discernment. The $m: 2_G^U$ in Definition \ref{defGET} indicates that  the $\phi$ is the focal elements outside of the frame of discernment, not the empty set in traditional BPA. Likewise, $m_G$ in Eq.(\ref{GBPA}) means GBPA assign some probability to the propositions, beyond the frame of discernment. For simplicity, the following of this paper without special explanation, BPA on behalf of GBPA, and the mass function $m_G$ simplifies  as $m$.

Similar to Dempster-Shafer theory, the generalized belief function (GBF) and generalized plausible function (GPL) in GET are defined as follows.
\begin{definition}\label{GBF}
Suppose a GBPA, the generalized belief function (GBF) is that GBel: ${2^U} \to [0,1]$, satisfied these conditions:
\begin{eqnarray}
GBel(A) = \sum\limits_{B \subseteq A} {m(B)} \\
GBel(\phi ) = m(\phi )
\end{eqnarray}
\end{definition}

\begin{definition}\label{GPF}
Suppose a GBPA, the generalized belief function (GPF) is that GPl: ${2^U} \to [0,1]$, satisfied these conditions:
\begin{eqnarray}
GPl(A) = \sum\limits_{B \cap A \ne \phi } {m(B)} \\
GPl(\phi ) = m(\phi )
\end{eqnarray}
\end{definition}

Note that in Definition \ref{GBF} and Definition \ref{GPF}, the values of $GBel(\phi )$ and $GPl(\phi )$ equal to $m(\phi )$, this is logical. Because $\phi$ is a proposition, beyond the fame of discernment, can not be supported  by these propositions within the frame of discernment, and also unknown whether agreed with these propositions beyond the frame of discernment. GBF and GPF can be regarded as  generalized lower bound and upper bound  in which subset $A$ has been supported, respectively. It is obviously that
\begin{equation}
GBel(A) \le GPl(A)
\end{equation}

\begin{example}
Suppose there is a frame of discernment of $\{a, b, c\}$, the GBPA is given as follows.
\[m\{ a\}  = 0.6;m\{ c\}  = 0.2;m\{ b,c\}  = 0.2\]
In this example, the GBPA assign just to these nonempty sets, that is $m(\phi)$=0. In this case, GBPA is the same to BPA.
\[\begin{array}{l}
GBel\{ a\}  = 0.6;GBel\{ b\}  = 0;GBel\{ c\}  = 0.2; GBel\{ b,c\}  = 0.2\\
GPl\{ a\}  = 0.6;Gpl\{ b\}  = 0.2;GPl\{ c\}  = 0.4; GPl\{ b,c\}  = 0.4
\end{array}\]
These results show that while  $m(\phi)$=0, the values of GBF and GPL in GBPA is the same to Bel and Pl in traditional BPA.
\end{example}

\begin{example}
Suppose there is a frame of discernment of $\{a, b, c\}$, the GBPA is given as follows.
\[m\{ a\}  = 0.6;m\{ b\}  = 0.1; m\{ b,c\}  = 0.2; m\{ \phi \}  = 0.1\]
In this example, the GBPA assign some value to the focal element $\phi$. The GBF and GPL are as follows.
\[\begin{array}{l}
GBel\{ a\}  = 0.6;GBel\{ b\}  = 0.1; GBel\{ c\}  = 0; GBel\{ b,c\}  = 0.2;GBel\{ \phi \}  = 0.1\\
GPl\{ a\}  = 0.6; GPl\{ b\}  = 0.3; GPl\{ c\}  = 0.2;GPl\{ b,c\}  = 0.3;GPl\{ \phi \}  = 0.1
\end{array}\]
\end{example}

\subsection{Generalized combination rule (GCR)}
The classic Dempster's combination rule can combine two BPAs $m_1$ and $m_2$ yield to a new BPA $m$. Based on the classic Dempster's combination rule, the generalized combination rule (GCR) is defined as follws.
\begin{definition}\label{GCR}
In generalized evidence theory,  ${\phi _1} \cap {\phi _2} = \phi$  means that the intersection between empty set and empty set is still an empty set.
Given two GBPA $m_1$ and $m_2$, the generalized combination rule (GCR) are defined as follows.
\begin{eqnarray}
&& m(A) = \frac{{(1 - m(\phi ))\sum\limits_{B \cap C = A} {{m_1}(B){m_2}(C)} }}{{1 - K}}\label{GCR1}\\
&& K = \sum\limits_{B \cap C = \phi } {{m_1}(B){m_2}(C)} \label{GCR2}\\
&& m(\phi ) = {m_1}(\phi ){m_2}(\phi )\label{GCR3}\\
&& m(\phi ) = 1 \quad If \quad and \quad only\quad  if \quad  K =1.\label{GCR4}
\end{eqnarray}
\end{definition}

The Characteristics of GCR, composed of Eqs.(\ref{GCR1}) -(\ref{GCR4}), are summarized as follows.\\
(1) When $m(\phi ) =0$, the GCR degenerate to Dempster's combination rule.\\
(2) The combination result of two empty sets can be obtained  by  multiplied  their  GBPAs values.\\
(3) The factor $1/(1 - K)$ in Eq.(\ref{GCR1}) is a normalized process that reassigned the GBPA values after deducting the $m(\phi)$ obtained from Eq.(\ref{GCR3}). In other words, the process is to multiply these GBPA, which intersections are not empty, and accumulate, at last amplify $1/(1 - K)$ times.\\
(4) Because of ${\phi _1} \cap {\phi _2} = \phi$, the GBPA of conflicting coefficient $K$ is obtained after superposing Eq.(\ref{DempsterK}) and Eq.(\ref{GCR3}).

There partial  properties of generalized evidence theory (GET) are shown as follows.
\begin{property}
When $m(\phi ) = 0 $, GBPA degenerates to traditional BPA. More generally, if GBPA just assigns in the single elements, GBPA degenerates to the probability of probability theory.
\end{property}

\begin{property}
For the GCR of GET, if  $m(\phi ) = 0$, then GCR degenerates to classic Dempster's combination rule. More generally, when GBPA just assigns in the single elements, the results of GCR is the same to Bayesian probability.
\end{property}

\begin{property}
The same as Dempster's combination rule, GCR satisfies commutativity and associativity. This is that the combination results by GCR is unrelated to the orders of combination.
\end{property}

\subsection{Generalized evidence distance}
The generalize evidence distance in GET is defined as follows.
\begin{definition}
Let $m_1$ and $m_2$ are two GBPAs on the frame discernment $\Theta$, the generalized evidence distance between GBPA $m_1$ and $m_2$ is defined as
\begin{equation}\label{gedistance}
{d_{GBPA}}({m_1},{m_2}) = \sqrt {\frac{1}{2}{{\left( {\overrightarrow {{m_1}}  - \overrightarrow {{m_2}} } \right)}^T}\overline D \left( {\overrightarrow {{m_1}}  - \overrightarrow {{m_2}} } \right)}
\end{equation}
where ${\overline D }$ is a ${2^N} \times {2^N}$ matrix, the elements of ${\overline D }$ is that
\begin{equation}
D(A,B) = \frac{{\left| {A \cap B} \right|}}{{\left| {A \cup B} \right|}};A,B \in P\left( \theta  \right)
\end{equation}
The detail calculation process is that
\[{d_{GBPA}}({m_1},{m_2}) = \sqrt {\frac{1}{2}\left( {{{\left\| {\overrightarrow {{m_1}} } \right\|}^2} + {{\left\| {\overrightarrow {{m_2}} } \right\|}^2} - 2\left\langle {\overrightarrow {{m_1}} ,\overrightarrow {{m_2}} } \right\rangle } \right)} \]
and ${\left\| {\overrightarrow m } \right\|^2} = \left\langle {\overrightarrow m ,\overrightarrow m } \right\rangle $,
$\left\langle {\overrightarrow m ,\overrightarrow m } \right\rangle $ is the inner product of the two vectors:

\[\left\langle {\overrightarrow {{m_1}} ,\overrightarrow {{m_2}} } \right\rangle  = \sum\limits_{i = 1}^{{2^N}} {\sum\limits_{j = 1}^{{2^N}} {{m_1}({A_i}){m_2}({A_j})} } \frac{{\left| {{A_i} \cap {B_i}} \right|}}{{\left| {{A_i} \cup {B_i}} \right|}};{A_i},{B_j} \in P(\theta )\]

\subsection{Application of GCR}

In this subsection, numerical examples are used to show the applications of GCR.
\begin{example}
Assume a frame of discernment $\Omega=(a,b,c)$, two GBPAs are given as:
\[\begin{array}{l}
{m_1}(a) = 0.5;{m_1}(a,b) = 0.5\\
{m_2}(a) = 0.5;{m_2}(b) = 0.3;{m_2}(\theta ) = 0.2
\end{array}\]

The process of calculation is that:
\[m(\phi ) = {m_1}(\phi ){m_2}(\phi ) = 0 \times 0 = 0\]
and
\[K = {m_1}(a){m_2}(b) = 0.5 \times 0.3 = 0.15\]
then
\begin{eqnarray*}
&& m(a) = \frac{{1 - m(\phi )}}{{1 - K}} \times (m_1(a)(m_2(a) + m_2(\theta )) + m_2(a)m_1(a,b)) \\
&& =\frac{{(1 - 0) \times (0.5 \times (0.5 + 0.2) + 0.5 \times 0.5)}}{{1 - 0.15}} = 0.706\\
&& m(b) = \frac{{1 - m(\phi )}}{{1 - K}} \times {m_1}(a,b){m_2}(b) = \frac{{(1 - 0) \times 0.5 \times 0.3}}{{1 - 0.15}} = 0.176\\
&&m(a,b) = \frac{{1 - m(\phi )}}{{1 - K}} \times {m_1}(a,b){m_2}(\theta ) = \frac{{(1 - 0) \times 0.5 \times 0.2}}{{1 - 0.15}} = 0.118\\
&&m(\theta ) = 0
\end{eqnarray*}
In this example, $m(\phi ) = 0$, the GCR is the same to the classic Dempster's combination rule.
\end{example}

\begin{example}
Suppose the frame of discernment  $\Omega=(a,b,c)$, two GBPAs are given as:
\[\begin{array}{l}
{m_1}(a) = 0.2;{m_1}(b) = 0.2;{m_1}(\phi ) = 0.6\\
{m_2}(a) = 0.2;{m_2}(b,c) = 0.1;{m_2}(\phi ) = 0.7
\end{array}\]
The conflicting coefficient $K$ in GET is calculated as:
\[\begin{array}{l}
K = {m_1}(a)({m_2}(b,c) + {m_2}(\phi )) + {m_1}(b)({m_2}(a) + {m_2}(\phi )) + {m_1}(\phi )({m_2}(a) + {m_2}(b,c) + {m_2}(\phi ))\\
 = 0.2 \times (0.1 + 0.7) + 0.2 \times (0.2 + 0.7) + 0.6 \times (0.2 + 0.1 + 0.7)\\ = 0.94
\end{array}\]
and
\[m(\phi ) = {m_1}(\phi ){m_2}(\phi ) = 0.6 \times 0.7 = 0.42\]
then
\[\begin{array}{l}
m(a) = \frac{{1 - m(\phi )}}{{1 - K}} \times {m_1}(a){m_2}(a) = \frac{{(1 - 0.42) \times 0.2 \times 0.2}}{{1 - 0.94}} = 0.387\\
m(b) = \frac{{1 - m(\phi )}}{{1 - K}} \times {m_1}(b){m_2}(b,c) = \frac{{(1 - 0.42) \times 0.2 \times 0.1}}{{1 - 0.94}} = 0.193\\
m(c) = 0
\end{array}\]
Thus, the final results are as follows
\[\begin{array}{l}
m(a) = 0.347;m(b) = 0.193;
m(c) = 0;m(\phi ) = 0.42
\end{array}\]
\end{example}

It is sure that, traditional Dempster's combination rule is not suitable for this situation, because of $m(\phi ) \neq 0$.
It is clearly that, after ascertained the value of $m(\phi)$, the rest probability is redistributed  to other nonempty sets, in GCR.
In this example, the probability of $m(c) $ is zero, because the single set $\{c\}$ is not supported by any one of the  two BPAs, in the frame of discernment. That is to say, the probability of single sets $\{a\}$ and $\{b\}$ are raised, owing to both of them are more or less supported by the two GBPAs.

\begin{example}\label{exampl5}
Suppose the frame of discernment  $\Omega=(a,b,c)$, two GBPAs are given as follows:
\[\begin{array}{l}
{m_1}(a) = 0.2;{m_1}(\phi ) = 0.8\\
{m_2}(b) = 0.5;{m_2}(\phi ) = 0.5
\end{array}\]
The conflicting coefficient $K$ in GET is calculated as:
\[\begin{array}{l}
K = {m_1}(a)({m_2}(b) + {m_2}(\phi )) + {m_1}(\phi )({m_2}(b) + {m_2}(\phi ))\\
 = 0.2 \times (0.5 + 0.5) + 0.8 \times (0.5 + 0.5)\\ = 1
\end{array}\]
thus,
\[m(\phi ) =1\]

From the view of GCR, we can first obtained the $m(\phi )={m_1}(\phi ) \times {m_2}(\phi )$ = 0.56.
However, since the rest two propositions are not supported by each other, thus the rest probability can not assign to any one of them, and reassigned to $m(\phi)$. Therefore, the $m(\phi)$ gets the twice assigned values, and $m(\phi )$ = 0.56 + 0.44 = 1. We believe that this  is reasonable. While the two GBPAs are highly conflicting and not supported by each others, the frame of discernment should be considered as an incomplete.
\end{example}

\begin{example}
Suppose the frame of discernment  $\Omega=(a,b,c)$, two GBPAs are given as follows:
\[\begin{array}{l}
{m_1}(a) = 0.65;{m_1}(b) = 0.35\\
{m_2}(\phi ) = 0.1
\end{array}\]
The result is that
\[m(\phi ) =1\]
This example is similar to Example \ref{exampl5}. Colloquially, $m(\phi )$ can be viewed as a number 0, and any numbers multiply 0 is still 0. If a GBPA assigns the total probability to the set $\phi$, the frame of discernment is incomplete.
\end{example}

\section{Application and Discussion}\label{SectAD}
\subsection{The $m(\phi)$ in Dempster-Shafer evidence and GET}
In classic Dempster-Shafer theory, $m(\phi)=0$ is indispensable. In view of the Dempster-Shafer theory, $\phi$ is a proposition without any supported by other propositions. That is non physical meaning of $\phi$ in Dempster-Shafer. The first observes $\phi$ and assigns the value to $m(\phi)$ is the TBM by Smets\cite{smets1994transferable}. In TBM, if the conflict between two bodies of evidence  is highly, the conflicting BPAs are assigned to $m(\phi)$.
Basic belief assignment (BBA) in TBM is used to distinguish the traditional BPA in Dempster-Shafer theory. However, the BBA and BPA are both assigned on these nonempty sets, essentially. In other words, the restriction condition of $m(\phi)=0$ is still necessary. Such an approach is questionable in  logic.
Now that the problem is in open world, TBM why not assign the value to $m(\phi)$ at the step of generate BBAs? In addition, the approach of dealing with conflict is humble. Yager proposed a method that when the evidences are highly conflicting, the conflicting values are assigned to the whole set $\Omega$\cite{yager1987dempster}. Yager's method is contentious in the following research. In GET, while the GBPAs are generated, $m(\phi) \neq 0$ is permissible. And this is easy to understand, that is there may exist some hypothesises or propositions beyond the fixed frame of discernment. The value of $m(\phi)$ indicates the open world degree of  frame of discernment.  Classic Dempster-Shafer theory is an extension of Bayes probability, and Dempster-Shafer theory can express and deal with uncertain information than Bayes probability, efficiently. And GET proposed in this paper, is an extension of classic Dempster-Shafer theory, and can express and deal with more uncertain information in the open world than Dempster-Shafer theroy in the close world.

\subsection{Modified Liu's conflict model}
As mentioned in Subsection \ref{liuconflict}, Liu analysed the shortage of classic conflict coefficient $K$, and proposed a conflict model. The main idea is that construct  a two tuples by pignistic betting distance $difBetP$ in TBM\cite{Smets1994191} and the classic conflict coefficient $K$ in Dempster-Shafer theory\cite{Dempster1967,Shafer1976}, and the union value to represent the conflict degree. That is when the value of  $difBetP$  is large, the two evidence can not be regarded as conflicting. While the value of $K$ is large, the two evidence can not be viewed as conflicting, also. Whether the evidences  is conflict just ascertained by  union  $difBetP$ with  $K$. The conflict model proposed is very interesting, since it provides a new thought of how to express the conflict between two bodies of evidence. There are some numerical examples by Liu's conflict model.
\begin{example}
Let a frame of discernment of $\theta  = \{ {\omega _1},{\omega _2},{\omega _3},{\omega _4},{\omega _5}\} $, and the two sensors $m_1$ and $m_2$ report that:
\[\begin{array}{l}
{m_1}({\omega _1}) = 0.8,{m_1}(\{ {\omega _2},{\omega _3},{\omega _4},{\omega _5}\} ) = 0.2\\
{m_2}(\theta ) = 1
\end{array}\]
The classic conflict coefficient  is that  $K = 0$. The pignistic betting distance between this two evidence is that $difBetP = 0.6 $. The two couples values of Liu's conflict model is that $c{f_{12}}(K,difBetP) = \left\langle {0,0.6} \right\rangle $, and can be regard as non conflict. This result is the same to Dempster-Shafer theory. However, it is clearly that the information of $m_1$ is richer than $m_2$. And the difference between the two evidence is given by  $difBetP = 0.6 $. That is the weights of two evidence is different.
\end{example}

\begin{example}
Let a frame of discernment of $\theta  = \{ a,b,c\}$, there are two BPAs as $m_1$ and $m_2$,
\[\begin{array}{l}
{m_1}(a) = \frac{1}{3};{m_1}(b) = \frac{1}{3};{m_1}(c) = \frac{1}{3}\\
{m_2}(a,b,c) = 1
\end{array}\]
Since $K= 0$ and $difBetP = 0$, $c{f_{12}}(K,difBetP) = \left\langle {0,0} \right\rangle $, and the two BPAs are regarded as not conflict. This is irrational obvious, because the two BPAs provide different information, and the $m_1$ is more definite, and $m_2$ is total ignorance.
\end{example}

From the above two examples, we know that "the same probability of occurrence" is the same to "the total ignorance of system" in Liu's conflict model.
However, the real situation is not so simple. When we known nothing about the system, that is $m(\Omega ) = 1$. It indicates that $m(a) = m(b) = 0.5$, $m(a) = 0.7; m(b) = 0.3$, $m(a) = 1; m(b) = 0$  and so forth are possible, and the probability of uncertainty of system is the biggest. Thus pignistic betting distance can not distinguish the same probability and total ignorance, and not be a well represent model of conflict. Based on Jousselme's evidence distance\cite{jousselme2001new}, we propose a modified evidence model of conflict.

\begin{definition}
Suppose there are two different BPAs on the same frame of discernment $\Theta$, the modified evidence conflict model is defined as follws
\begin{equation}\label{modifiedLiu}
cf({m_1},{m_2}) = \left\langle {K,dis} \right\rangle
\end{equation}
where $K$ is the classic conflict coefficient in Dempster-Shafer evidence by Eq.(\ref{DempsterK}), and $dis$ is the evidence distance by Eq.(\ref{Jousselmedistance})
\end{definition}

\subsection{Conflict model of GET}
In the previous subsection, we modify Liu's conflict model by Eq.(\ref{modifiedLiu}), and it is more reasonable. However, the conflict model is still in the close world, and can not be applied to the open world, which frame of discernment may be incomplete. In view of this, based on GET, the new conflict model of GET is proposed.
\begin{definition}
Assume there are two GBPAs $m_1$ and $m_2$ on the frame of discernment $\Theta$, then the generalized conflict model is given as
\begin{equation}\label{gcm}
c{f_G}({m_1},{m_2}) = \left\langle {K,dis} \right\rangle
\end{equation}
\end{definition}
Where $K$ is the generalized conflict coefficient by Eq.(\ref{GCR2}), and $dis$ is the generalized evidence distance by Eq.(\ref{gedistance}).
\end{definition}

Compared with these existing methods, this new proposed generalized conflict model can handle with the information on the incomplete frame of discernment.
When the frame of discernment is complete, the generalized conflict coefficient $K$ degenerates to classical coefficient $K$ in Dempster-Shafer theoryEq.(\ref{DempsterK}), and the Eq.(\ref{gcm}) degenerates to  Eq.(\ref{modifiedLiu})

The below examples are used to show the application of generalized conflict coefficient and generalized conflict model.
\begin{example}
There are to GBPAs $m_1$ and $m_2$ on the frame of discernment. The first group GBPAs is varied, begin with  ${m_1}(a) = 1$ and $m_1(\phi ) = 0$, and ${m_1}(a)$ decreases progressively 0.1 each step , $m(\phi )0$ progressively increases 0.1 each step . The second group GBPAs is also varied, begin with  ${m_2}(a) = 1$ and $m_2(\phi ) = 0$, and ${m_2}(a)$ decreases progressively 0.1 each step , $m_1(\phi )0$ progressively increases 0.1 each step . Then the generalized conflict coefficient $K$ between the two GBPAs is shown in Fig.\ref{generalizedconflictcoefficient}.

\begin{figure}[htbp]
\centerline{\psfig{file=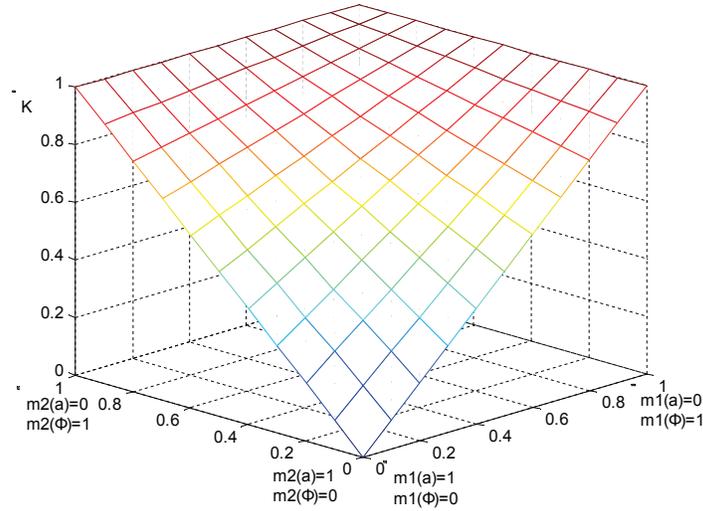,scale=0.6}}
\vspace*{8pt}
\caption{Generalized conflict coefficient between two GBPAs.}
\label{generalizedconflictcoefficient}
\end{figure}
Fig.\ref{generalizedconflictcoefficient} indicates that, when ${m_1}(a) = 1$; $m_1(\phi ) = 0$ and ${m_2}(a) = 1$; $m_2(\phi ) = 0$, the generalized conflict coefficient is the smallest, and the proposition $\{a\}$ is absolutely supported by the system and  the frame of discernment is complete.
While  ${m_2}(a)$ decreases progressively 0.1 each step, $m_1(\phi )0$ progressively increases 0.1 each step, the generalized conflict coefficient is also gradually increased, and this situation indicates the incomplete degree of the frame of  discernment is bigger and bigger. According to GCR, when $m(\phi ) = 1$ appeared in anyone of these GBPAs, the generalized conflict coefficient achieves the  maximum value  1.
\end{example}

\begin{example}
Suppose there are two GBPAs on the frame of discernment. The first group GBPAs is varied, begin with  ${m_1}(a) = 0$ and $m_1(\phi ) = 1$, and ${m_1}(a)$ progressively increases 0.1 each step, $m(\phi )0$  decreases progressively 0.1 each step. The second group GBPAs is also varied, begin with  ${m_2}(a) = 0 $ and $m_2(\phi ) = 1 $, and ${m_2}(a)$ progressively increases 0.1 each step , $m_2(\phi ) $ decreases progressively 0.1 each step. Then the generalized evidence distance  between the two GBPAs is shown in Fig.\ref{generalizedevidencedistance}.
\begin{figure}[htbp]
\centerline{\psfig{file=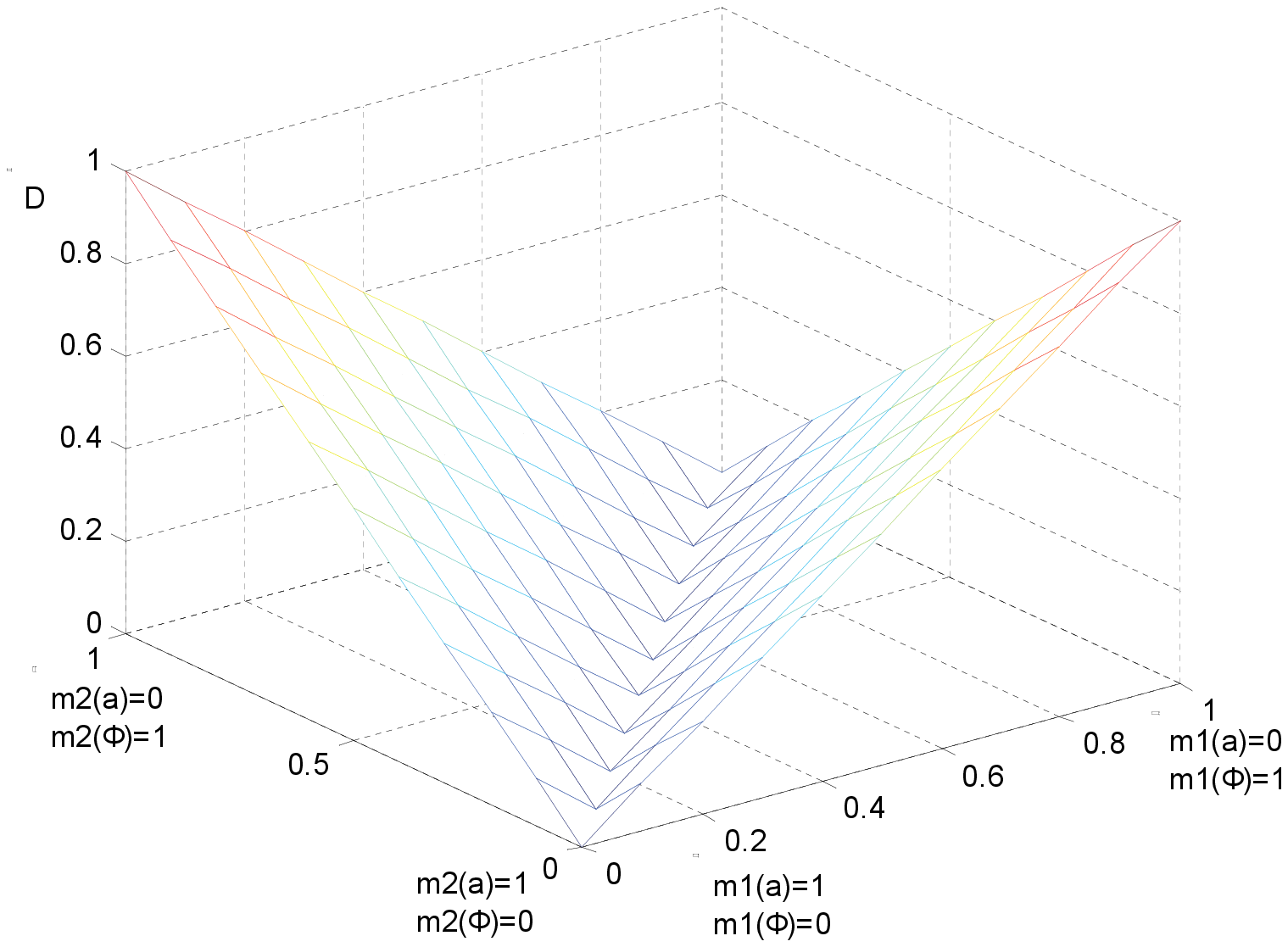,scale=0.6}}
\vspace*{8pt}
\caption{Generalized evidence distance between two GBPAs.}
\label{generalizedevidencedistance}
\end{figure}
From Fig.\ref{generalizedevidencedistance}, we can see that the generalized evidence distance on axis keeps 0, because the two GBPs are the same.
\end{example}

\begin{example}
Let two GBPAs on the same discernment, and the two GBPAs are given as
\[\begin{array}{l}
{m_{x1}}(b) = 0.1;{m_{x1}}(\phi ) = 0.9\\
{m_{x2}}(b) = 0.1;{m_{x2}}(\phi ) = 0.9
\end{array}\]

The value of conflict model of GET can be obtained as follows
\[c{f_G}({m_1},{m_2}) = \left\langle {0.81,0} \right\rangle \]
and can be seen in Fig.\ref{twosamplewithdifferentclasses}
\begin{figure}[htbp]
\centerline{\psfig{file=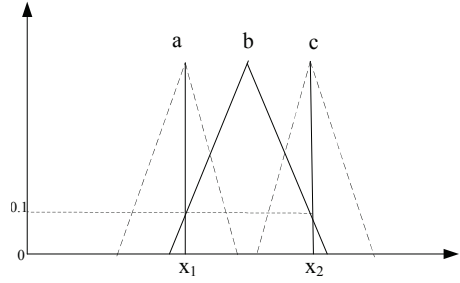,scale=0.7}}
\vspace*{8pt}
\caption{Two samples with different classes.}
\label{twosamplewithdifferentclasses}
\end{figure}
Fig.\ref{twosamplewithdifferentclasses} indicates that there is just one class  $b$ on the frame of discernment. The two dotted triangles represent the classes a and c, respectively, and both of them do not appeared on this frame of discernment. In fact, the sensors $x1$ and $x2$ belong to classes $a$ and $c$, respectively. If generalized evidence distance is considered  merely, the error result will be obtained.  This example points out that generalized conflict coefficient to measure the  conflict is better than generalized evidence distance on the incomplete frame of discernment.
 \end{example}
The two below examples are applied to the complete frame of discernment with $m(\phi)= 0$.

\begin{example}
Suppose a frame of discernment $\theta  = \{ a,b\} $, the first fixed BPA is that: $m_1(a,b) = 1$.
The second varied BPA  begin with $m_2(a) = 0; m_2(b) = 1$.  And ${m_2}(a)$ progressively increases 0.1 each step , $m_2(b)$ decreases progressively 0.1 each step. Then the evidence distance and pignistic betting distance between $m_1$ and $m_2$ are shown in Fig.\ref{evidencesvsppt}.
\begin{figure}[htbp]
\centerline{\psfig{file=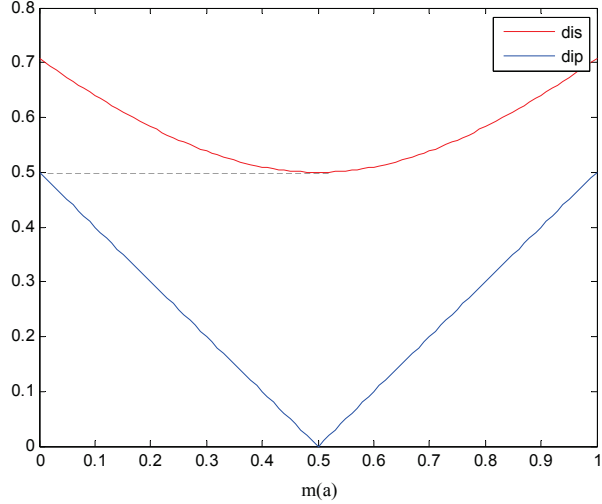,scale=0.6}}
\vspace*{8pt}
\caption{Comparisons with evidence distance and pignistic betting distance.}
\label{evidencesvsppt}
\end{figure}
From Fig.\ref{evidencesvsppt}, When $m_2(a) = 0.5; m_2(b) = 0.5$, the pignistic betting distance between two BPAs is 0, which indicates no conflict between BPAs. However, at this time step, the value of evidence distance is 0.5, which indicates that there exists conflict between BPAs.
It is obviously that, when the frame of discernment is complete, the measure of conflict model should be  evidence distance but not the pignistic betting distance.
\end{example}

\begin{example}
Let a frame of discernment of $\Omega=\{1,2,3  \cdots   20\}$, and the two BPAs are defined as follows
\[\begin{array}{l}
{m_1}(\{ 7\} ) = 0.6;{m_1}(A) = 0.4\\
m2(\{ 1,2,3\} ) = 1
\end{array}\]
where $A$ is a varied subset of $\Omega$, which increments one more element at a time, and begins with A=$\{1\}$, ending with Case 20 when $A = \{1,2,3  \cdots   20\}$.
The evidence distance and classic conflict coefficient $K$ between two BPAs are given in Table \ref{dbpavsk} and Fig.\ref{dbpavskfig}.
\begin{table}[!htbp]
\caption{Comparison of new proposed conflict coefficient with classic conflict coefficient.}
	\label{dbpavsk}
    \begin{tabular}{lll}\hline
    Cases       & $d_{BPA}$ & $K$   \\\hline
    A=\{1\}     & 0.7916 & 0.6 \\
    A=\{1,2\}   & 0.7024 & 0.6 \\
    A=\{1,2,3\} & 0.6    & 0.6 \\
    A=\{1,...,4\} & 0.6782 & 0.6 \\
    A=\{1,...,5\} & 0.7211 & 0.6 \\
    A=\{1,...,6\} & 0.7483 & 0.6 \\
    A=\{1,...,7\} & 0.7982 & 0.6 \\
    A=\{1,...,8\} & 0.8    & 0.6 \\
    A=\{1,...,9\} & 0.8083 & 0.6 \\
    A=\{1,...,10\} & 0.8149 & 0.6 \\
    A=\{1,...,11\} & 0.8202 & 0.6 \\
    A=\{1,...,12\} & 0.8246 & 0.6 \\
    A=\{1,...,13\} & 0.8283 & 0.6 \\
    A=\{1,...,14\} & 0.8315 & 0.6 \\
    A=\{1,...,15\} & 0.8343 & 0.6 \\
    A=\{1,...,16\} & 0.8367 & 0.6 \\
    A=\{1,...,17\} & 0.8388 & 0.6 \\
    A=\{1,...,18\} & 0.8406 & 0.6 \\
    A=\{1,...,19\} & 0.8423 & 0.6 \\
    A=\{1,...,20\} & 0.8438 & 0.6 \\\hline
    \end{tabular}
\end{table}

\begin{figure}[!htbp]
\centerline{\psfig{file=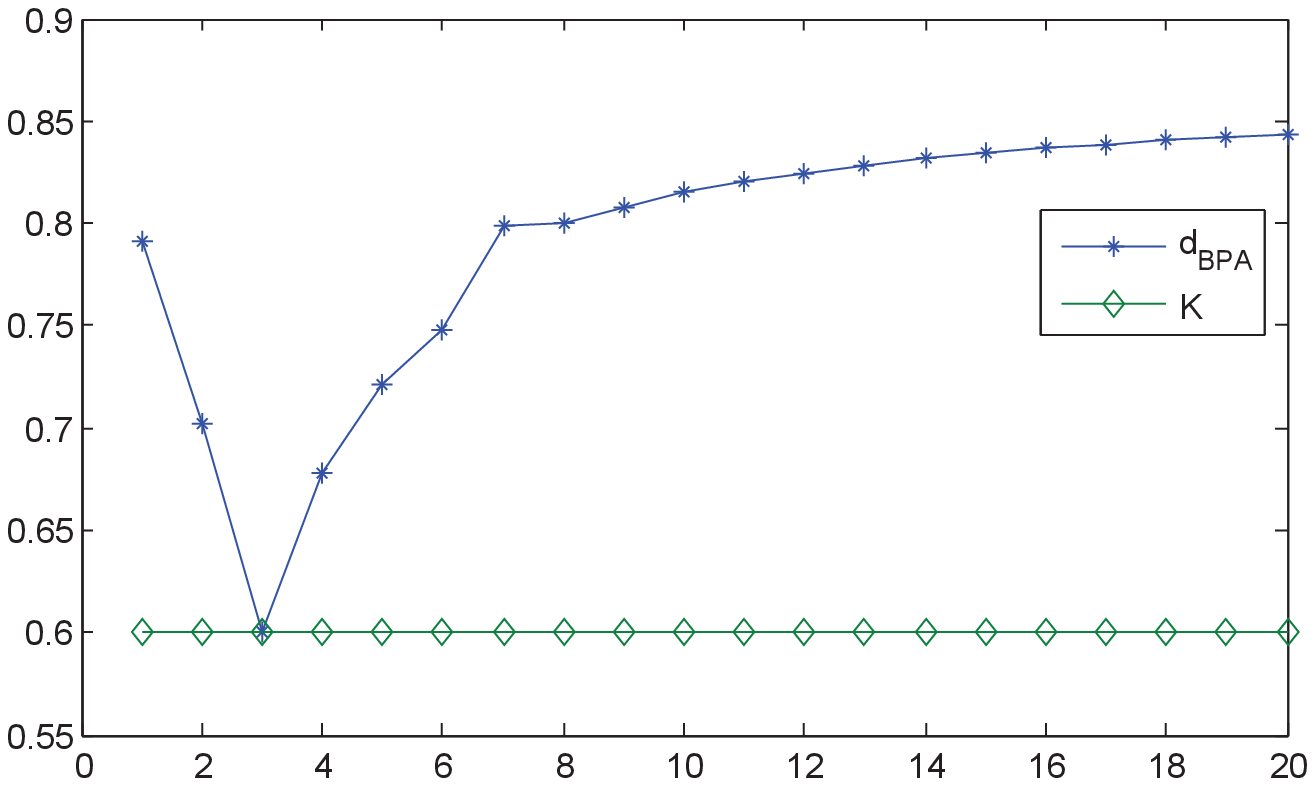,scale=0.6}}
\vspace*{8pt}
\caption{Comparison of new proposed conflict coefficient with classic conflict coefficient.}
\label{dbpavskfig}
\end{figure}
From Table \ref{dbpavsk} and Fig.\ref{dbpavskfig}, we can see that, no matter how the subset $A$ changed, the classic conflict coefficient $K$ keeps the value 0.6. This is irrational. The generalized evidence distance is  changed along with subset $A$, and it can measure the conflict efficiently. That is the the generalized evidence distance to measure the conflict is better than generalized conflict coefficient in the case of complete frame of discernment.
\end{example}

These numerical examples indicate the applications  of the new conflict model. When the frame of discernment is incomplete and $m(\phi) \ne  0$, the two-tuples of conflict model to measure the conflict should depend on generalized conflict coefficient mainly. On the other hand, when the frame of discernment is complete and $m(\phi) =  0$, the two-tuples of conflict model to measure the conflict should depend on generalized evidence distance mainly.

\section{Conclusions}\label{SectConclusions}
In this paper, a novel theory called generalized evidence theory is systematically proposed. Some key points can be concluded as follows.

First, evidence theory is an efficient tool to handle the information fusion under uncertain environment.Under the condition that we definitely know that we are in close world, the Dempster combination rule is enough to combine the evidence from different sources. If the evidence conflicts with each other, with more and more evidence are collected, the conflict evidence combination can be well addressed by taking the reliability of each evidence into consideration.

Second, generalized evidence theory (GET)is an extension of Dempster-Shafer theory, since the strict restriction condition of $m(\phi) = 0$ is abandoned.
In GET, $\phi$ is regards as a element, which means unknown but not just empty, with all of the property of other elements. That is GET can deal with more uncertain information fusion in the open world than Dempster-Shafer theory.
GET inherits all the merits of Dempster-Shafer theory, and expands the application of Dempster-Shafer theory, from close world to open world. When the frame of discernment is complete, and  $m(\phi) = 0$, GET degenerates to Dempster-Shafer theory.
And the generalized combination rule (GCR) is provided in GET, the distinction between GCR and Dempster combination rule is  $m(\phi)$.

Third, the generalized conflict model is given based on GET. Compared with these existing conflict model, the new conflict model is generalized. Under different conditions, how to apply the generalized conflict model is provided.  That is when the frame of discernment is incomplete, the measure of generalized conflict should take advantage of generalized conflict coefficient in main. When the frame of discernment is complete, the measure of generalized conflict should take advantage of evidence distance in main.

It should be pointed out that, besides the conflict management in open world, the exclusive condition in Dempster Shafer evidence theory should also be paid great attention. One of the ongoing works is a new theory called D numbers theory which focuses on the mutual exclusion in evidence theory. In classical evidence theory, it is assumed that each hypothesis is exclusive to each other. However, it is not reasonable in real world. For example, given two linguistic values Good and Very Good, the $m{G,VG}=0.8$ is not accepted under the framework of Dempster Shafer evidence theory since that the two linguistic values are not exclusive with each other. To address this limitation, a novel theory called D numbers theory is proposed \cite{Deng2012}. Both the D numbers theory and the proposed GET in this paper are generalization of the evidence theory, providing a more flexible and reasonable way to handle the uncertainty in our real world \cite{deng2014supplier,deng2014environmental,liu2014failure}.

\section*{Acknowledgements}
The author greatly appreciates Professor Shan Zhong, the China academician of the Academy of Engineering, for his encouragement to do this research. The author also greatly appreciates Professor Yugeng Xi in Shanghai Jiao Tong University for his support to this work. Professor Sankaran Mahadevan in Vanderbilt University discussed many relative topics about this work. Dr. Deqiang Han and Dr. Wen Jiang discussed the topic of conflict management in this paper. My Ph.D students in Shanghai Jiao Tong University, Peida XU and Xiaoyan Su, do some numerical experiments of this work. The author's Ph.D students in Southwest University, Xinyang Deng, Ya Li and Daijun Wei, graduate students in Southwest University Yajuan Zhang, Bingyi Kang, Xiaoge Zhang, Shiyu Chen, Yuxian Du, Cai Gao, have discussed the conflict evidence management. Mr. Hongming mo did some editorial works of this submission. This work is partially supported by National Natural Science Foundation of China, Grant Nos. 30400067, 60874105, 60904099, 61174022, Chongqing Natural Science Foundation, Grant No. CSCT, 2010BA2003, Program for New Century Excellent Talents in University, Grant No.NCET-08-0345, Shanghai Rising-Star Program Grant No.09QA1402900, the Chenxing Scholarship Youth Found of Shanghai Jiao Tong University, Grant No.T241460612, Doctor Funding of Southwest University Grant No. SWU110021, the open funding project of State Key Laboratory of Virtual Reality Technology and Systems, Beihang University (Grant No.BUAA-VR-14KF-02).

\bibliographystyle{elsarticle-num}
\bibliography{references}

\end{document}